\documentclass[10pt,twocolumn,letterpaper]{article}

\usepackage[pagenumbers]{cvpr} 

\usepackage[dvipsnames]{xcolor}
\usepackage{float}
\usepackage{dcolumn}
\usepackage{acronym}
\usepackage{multirow}
\usepackage{listings}
\usepackage{times}
\usepackage{epsfig}
\usepackage{graphicx}
\usepackage{amsmath}
\usepackage{amssymb}
\usepackage{xspace}
\usepackage{booktabs}
\usepackage{microtype}
\usepackage{multirow}
\usepackage[dvipsnames]{xcolor}
\usepackage[leftcaption]{sidecap}
\usepackage{enumitem}
\usepackage{rotating}
\usepackage{dblfloatfix}
\usepackage{listings}
\usepackage{float}

\definecolor{codegreen}{rgb}{0,0.6,0}
\definecolor{codegray}{rgb}{0.5,0.5,0.5}
\definecolor{codepurple}{rgb}{0.58,0,0.82}
\definecolor{backcolour}{rgb}{0.95,0.95,0.92}
\lstdefinestyle{mystyle}{
    backgroundcolor=\color{backcolour},   
    commentstyle=\color{codegreen},
    keywordstyle=\color{magenta},
    numberstyle=\tiny\color{codegray},
    stringstyle=\color{codepurple},
    basicstyle=\ttfamily\scriptsize,
    breakatwhitespace=false,         
    breaklines=true,                 
    captionpos=b,                    
    keepspaces=true,                 
    numbers=left,                    
    numbersep=5pt,                  
    showspaces=false,                
    showstringspaces=false,
    showtabs=false,                  
    tabsize=2,
    showlines=true
}

\newcommand{\methodname}[1]{ProViQ}
\newcolumntype{d}[1]{D{(}{(}{#1}}

\usepackage[dvipsnames]{xcolor}
\usepackage{float}
\definecolor{cvprblue}{rgb}{0.21,0.49,0.74}
\usepackage[pagebackref,breaklinks,colorlinks,citecolor=cvprblue]{hyperref}
\usepackage{acronym}

\def\paperID{12418} 
\def\confName{CVPR}
\def\confYear{2024}

\title{Zero-Shot Video Question Answering with Procedural Programs}

\author{Rohan Choudhury$^1$ \quad Koichiro Niinuma$^2$ \quad Kris M. Kitani$^1$
\quad L\'{a}szl\'{o} A. Jeni$^1$
\\
${^1}$Robotics Institute, Carnegie Mellon University \quad ${^2}$Fujitsu Research of America\\
{\tt\small {\{rchoudhu, kmkitani\}@andrew.cmu.edu}} \quad {\tt\small kniinuma@fujitsu.com}\quad {\tt\small laszlojeni@cmu.edu} \\
}

\begin{document}
 \vspace{-0.5em}
 \twocolumn[{%
 \renewcommand\twocolumn[1][]{#1}%
 \maketitle
 \begin{center}
 \centering
 \captionsetup{type=figure}
 \resizebox{\textwidth}{!}{\includegraphics[width=1.03\linewidth, height=6cm]{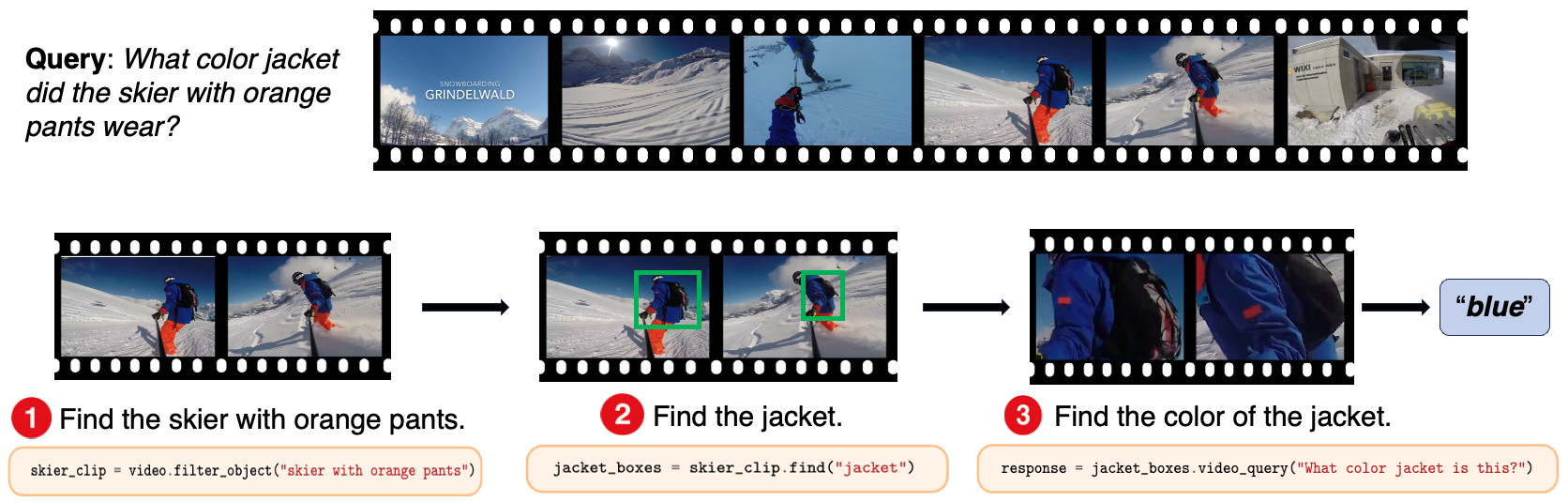}}
 \captionof{figure}{Our method, \textbf{\methodname\.}, reasons procedurally about videos by generating and executing Python programs that solve visual subtasks, mimicking how humans might approach such problems.}
 \label{fig:teaser}
 \end{center}
 }]

\begin{abstract}
\vspace{-1.5em}

We propose to answer zero-shot questions about videos by generating short procedural programs that derive a final answer from solving a sequence of visual subtasks. We present \textbf{Pro}cedural \textbf{Vi}deo \textbf{Q}uerying (\methodname\.), which uses a large language model to generate such programs from an input question and an API of visual modules in the prompt, then executes them to obtain the output. Recent similar procedural approaches have proven successful for image question answering, but videos remain challenging: we provide \methodname\. with modules intended for video understanding, allowing it to generalize to a wide variety of videos. This code generation framework additionally enables \methodname\. to perform other video tasks in addition to question answering, such as multi-object tracking or basic video editing. \methodname\. achieves state-of-the-art results on a diverse range of benchmarks, with improvements of up to 25\% on short, long, open-ended, and multimodal video question-answering datasets. Our project page is at \href{https://rccchoudhury.github.io/proviq2023}{this url}.
\end{abstract}    
 \vspace{-1.5em}
\section{Introduction}
\label{sec:intro}
Consider the video in Figure 1. What color jacket did the skier in orange pants wear? To answer this question, one might look for a skier in each frame, search those frames for one with orange pants, then check what color jacket they were wearing. Humans tend to solve such questions \textit{procedurally}, breaking problems down into a sequence of steps, each with concrete results. We hypothesize that using this type of reasoning can significantly improve performance for zero-shot video question answering (QA).

Existing video QA methods do not follow this approach. The predominant paradigm for video understanding is to train a supervised end-to-end model, typically by pre-training on large video datasets such as Kinetics \cite{carreira2017quo}  or Ego4D \cite{grauman2022ego4d}, then fine-tuning on relatively smaller QA benchmarks. Some zero-shot QA methods such as FrozenBilM \cite{yang2022frozenbilm} combine pre-trained video backbones and language models with some success, but are still unable to explicitly carry out procedural reasoning. However, more recently ViperGPT\cite{Suris_2023_ICCV} and VISPROG\cite{gupta2023visual} used large language models (LLMs) to generate short programs for this exact type of reasoning, achieving strong results for compositional image tasks. ViperGPT in particular demonstrated promising results on the NeXT-QA video dataset, but was limited by its image-centric approach from generalizing to other diverse video benchmarks.

Inspired by these works, we propose \textbf{Pro}cedural \textbf{Vi}deo \textbf{Q}uerying (\methodname\.), which uses an LLM to generate Python programs to answer zero-shot \textit{video} queries. Recent LLMs like ChatGPT and GPT4\cite{openai2023gpt} have demonstrated tremendous ability to generate high-quality code. \methodname\. takes advantage of this, providing the LLM with a prompt containing an API of visual modules for video reasoning, which it can use as steps in the generated program. For example, it can use a retrieval module to find all the frames with people in them, an object detector to find yogurt, and an image QA module to check what utensil was being used. We include modules for image and video-based reasoning, namely object detection, image QA, video retrieval, video captioning, speech transcription, tracking, and video summarizing, as part of the provided API for the program. As a result, \methodname\. can reason at multiple semantic levels by considering information from individual frames, disjoint video clips, and the whole video.

Using procedural reasoning in this manner has several additional benefits. Firstly, \methodname\. requires no further training. LLMs can already produce working Python programs, and the provided modules use models pre-trained on large image and video datasets, enabling them to generalize well when used for simple tasks.
This allows smooth incorporation of task-specific modules, making it simple to add  capabilities for solving new types of questions without fine tuning.
Secondly, the program's reasoning is interpretable: each line in the program can help attribute errors to specific modules. Thirdly, the LLM can compose the modules freely, enabling capabilities beyond question-answering. For example, combining the object detector and tracker modules yields a query-based multi-object tracking system, and the retrieval module can be used for basic video editing, just through generating a few lines of code.

\methodname\. leverages these advantages to significantly improve on a wide range of zero-shot video question answering benchmarks: we improve up to 25$\%$ on ActivityNet \cite{yu2019activityqa}, iVQA \cite{yang2021justask}, MSR-VTT-QA \cite{xu2017video}, MSVD-QA \cite{xu2017video}, TGIF-QA \cite{jang2017tgif} and NeXT-QA \cite{xiao2021next}, \textit{without any additional training}, even surpassing the \textit{supervised} state-of-the-art on some datasets. We also demonstrate strong performance on understanding long egocentric videos with a gain of $25\%$ on the challenging EgoSchema benchmark \cite{mangalam2023egoschema}, as well as multimodal understanding, obtaining state-of-the-art performance on the TVQA dataset \cite{lei2018tvqa}.

In summary, our contributions are that
\begin{enumerate}
    \item We present \methodname\. , a method to procedurally reason about and answer zero-shot video queries through generated Python programs.
    \item We show \methodname\.'s flexibility for tasks beyond question answering, such as query-based multi-object tracking and video editing.
    \item We achieve large accuracy improvements on a wide range of video QA benchmarks, and provide extensive ablations to demonstrate the utility of our approach.
\end{enumerate}

 \section{Related Work}
\label{sec:related_works}
\begin{figure*}[!htbp]
    \centering
    \includegraphics[width=0.9\textwidth, height=8cm]{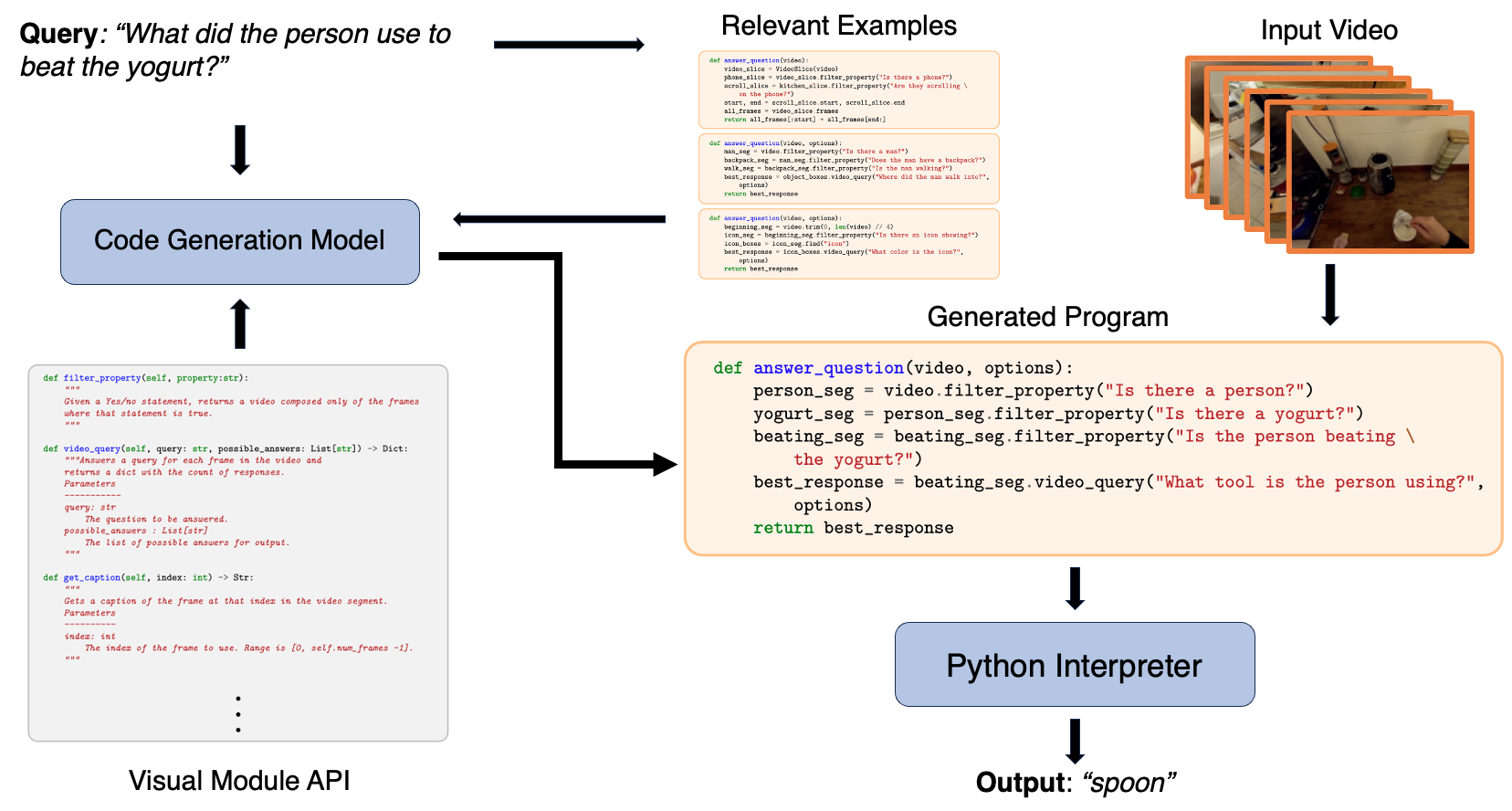}
    \caption{\textbf{Our method.} \methodname\. provides the input question, visual API, and relevant examples in the prompt to the code generation model. The LLM generates a short Python program that uses modules from the API to answer the question. Using the video as input, we execute the program to obtain the final output.}
    \label{fig:enter-label}
\end{figure*}

\subsection{Video Question Answering}
Compared to other video tasks, video question-answering datasets are relatively small. As in image QA, the predominant paradigm for video QA is to pre-train models on large datasets like Kinetics \cite{carreira2017quo}, Ego4D \cite{grauman2022ego4d}, HowTo100M \cite{miech2019howto100m}, or YouTube-100M \cite{zellers2022merlot}, then fine-tune on smaller annotated QA datasets \cite{fan2019heterogeneous, gao2018motion, huang2020location, seo2021look, park2021bridge, lin2021vx2text, le2020hierarchical, kim2020modality, fu2021violet, ye2023hitea}. Recent work like InternVideo \cite{wang2022internvideo} and mPLUG-Owl \cite{ye2023mplug} scale up this type of training significantly, but still do not generalize well enough to answer questions about videos outside their training distribution zero-shot.

Comparatively few works directly address zero-shot video QA.  BLIP \cite{li2022blip} trains a large model on image-question-answer triplets and evaluates the transfer to video QA tasks, but includes a fine-tuning step. Current methods are typically trained on web-scale datasets with audio or speech transcripts providing weak language supervision \cite{yang2021justask, zellers2021merlot, zellers2022merlot, yang2022frozenbilm}. In particular, FrozenBiLM \cite{yang2022frozenbilm} connects a frozen bidirectional language encoder with a trainable video model, trains on WebVid10M \cite{bain2021frozen} and measures zero-shot question-answering performance. These methods currently obtain state-of-the-art results, and we compare against them in Section \ref{sec:zeroshot_vqa}. Another recent line of work \cite{2023videochat, song2023moviechat, Maaz2023VideoChatGPT} use combinations of LLMs and visual inputs such as textual descriptions or CLIP \cite{radford2021learning} features from sparsely sampled frames to enable conversations about videos, but do not achieve strong quantitative results on standard benchmarks.

\subsection{Modular Vision}
Neural Modular Networks (NMNs) \cite{andreas2016neural} introduced modular visual question answering approaches, using parsers to compose learned modules into single trainable network. Follow-up methods to NMNs jointly trained the layout generator and the visual modules with reinforcement learning and weak supervision \cite{hu2017learning, hu2018explainable, saqur2020multimodal}. Several other works train large models that contain modules for different modalities and tasks \cite{zeng2022socratic, girdhar2023imagebind, reddy2022mumuqa,  wang2022language}, but cannot freely compose them or alter their layout.

In the past year, CodeVQA \cite{subramanian-etal-2023-modular}, VISPROG \cite{gupta2023visual} and ViperGPT \cite{Suris_2023_ICCV} leveraged the large improvements in language modeling to reformulate modular VQA as code generation: they use the strong performance of GPT-3 \cite{brown2020language} and GPT-4\cite{openai2023gpt} on code generation to formulate answers to visual questions as short Python programs, enabling use of mathematical operations, if-statements, and logical operators to manipulate the outputs of visual models. ViperGPT provides strong results on the NeXT-QA dataset \cite{xiao2021next}, but cites the length and inability to temporally reason as limiting factors for further experiments. We directly build off ViperGPT's approach with the same program generation framework and successfully extend it to video QA.

\subsection{Prompting and Tool Use}
With the recent surge of interest in large language models, many papers have studied how to effectively incorporate additional tools, either through fine-tuning \cite{schick2023toolformer} or prompting \cite{wei2022chain, yang2021empirical, 2023videochat, yang2023gpt4tools}. Following the success of large multimodal models like Flamingo \cite{alayrac2022flamingo} and GPT-4 \cite{openai2023gpt}, recent methods train multimodal language models, such as LLAVA \cite{liu2023llava, liu2023improvedllava},  X-GPT \cite{zou2023generalized} or MiniGPT-4  \cite{zhu2023minigpt} to add visual capabilities to language models. However, these have not successfully incorporated videos, due to the challenges of training on large-scale video data. 

Following CodeVQA, VISPROG and ViperGPT, we use several pre-trained models for visual functions like object detection and image QA. We use GroundingDINO \cite{liu2023grounding} for text-conditioned object detection, and BLIP-2 \cite{li2023blip} for image-conditioned captioning and QA. We also use LaViLa \cite{zhao2023learning} for video-to-language generation and use GPT3.5 \cite{brown2020language} for generating code and querying summaries. We use ByteTrack \cite{zhang2022bytetrack} for multi-object tracking.

\section{Method}
\begin{figure*}[!htbp]
    \centering
    \includegraphics[width=\textwidth, height=8cm]{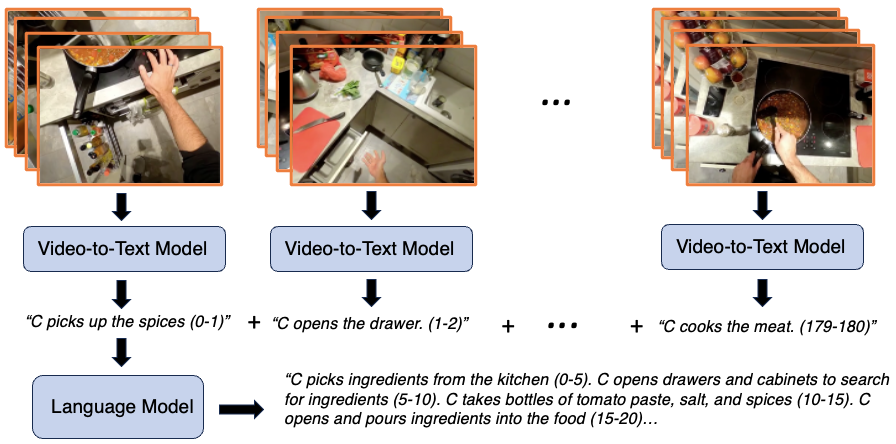}
    \caption{\textbf{Our long video summarization module.} In order to answer questions about video narratives, our summarization module uses an LLM to fuse the outputs of video-to-text models on subclips into a coherent paragraph summary. This method produces qualitatively accurate summaries, but depends on a high-quality video-to-text model that captures the action in short video clips. We used LaViLa \cite{zhao2023learning}, which is meant to be used on the Ego4D dataset.}
    \label{fig:long_summarization}
\end{figure*}
\subsection{Program Generation}

The input to \methodname\. is an input video and a query. We then construct a prompt containing a list of video modules in an application programming interface (API), the input query or task, and a few example programs, which is then fed to the LLM (we used \texttt{gpt-3.5-turbo} for all experiments). The LLM produces a short Python program that decomposes the input query into concrete steps, each calling visual modules specified by the API. The generated program takes as input the question, the video frames, and other relevant information such as a list of multiple choice options. Once the output is generated, we compile and execute the program as in \cite{Suris_2023_ICCV} using Python's built-in \texttt{exec()} function. The compiled function runs on the input video to output the final answer or modified video as specified by the task.

\subsection{Video Modules}

The list of video modules in the provided API are intended as a toolbox for the generated programs to use for decomposing and answering questions. We used these specific methods in order to encompass a wide variety of datasets and possible questions, with room for the LLM to improvise and combine methods together as it sees fit. While this API is not necessarily exhaustive, we found it sufficient to answer the vast majority of questions in QA benchmarks. We provide this list of modules to the LLM in the prompt in the form of methods with documentation, which we include in Appendix \textcolor{red}{E}. While all modules are intended to run on collections of frames, they work on single images as well, and thus enable reasoning about singular frames as well as clips. Specifically, they include:

\noindent\textbf{\texttt{filter\_property}} Given a boolean predicate such as \textit{``Is the person running?"}, this method finds all frames in the video that satisfy the predicate. It works by calling BLIP-2 on an input batch of frames with the question and collecting all frames where the answer is \texttt{yes}. 

\noindent\textbf{\texttt{filter\_object}} Given a specific object, such as \textit{yogurt}, this method fnids all frames in the video where the object exists.This method runs an object detector over the input clip and returns all the frames where the given object is found. While \texttt{filter\_property} can handle this functionality, using an object detector works much better for finding specific objects.
\begin{figure*}[!htbp]
    \centering
    \includegraphics[width=0.9\textwidth, height=10cm]{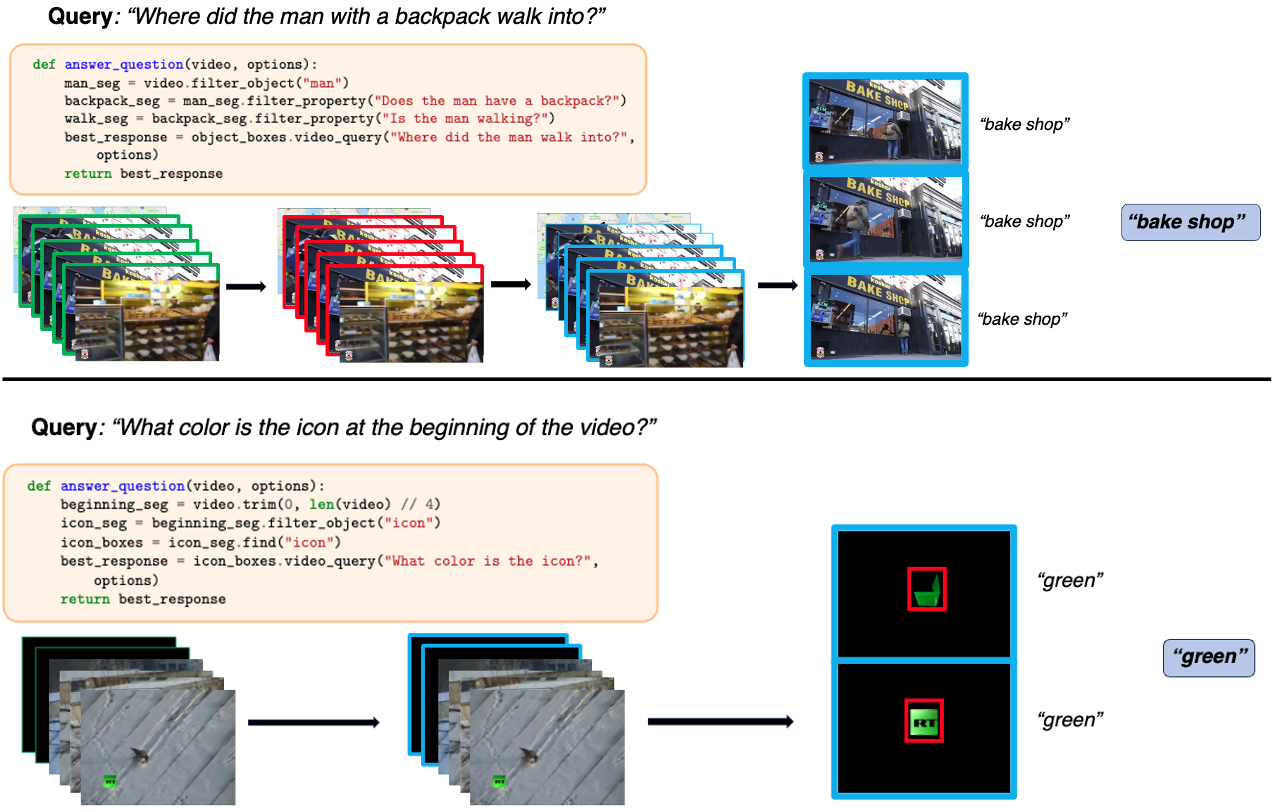}
    \caption{\textbf{Qualitative samples from \methodname\. .} Both samples demonstrate how \methodname\. can decompose a query into steps, translate those into function calls, and execute them to procedurally obtain an answer. It combines the use of object detectors, retrieval methods, captioning and image QA, among other tools, to solve complex questions zero-shot.}
    \label{fig:vidprog_samples}
\end{figure*}

\noindent\textbf{\texttt{find}} This method calls a text-conditioned object detector to return each crop of of the input object found in the collection of frames. This method is useful for zooming in to the input object over time, which can improve performance on visual queries about it.

\noindent\textbf{\texttt{video\_query}} This method invokes BLIP-2 \cite{li2023blip}'s QA capability to answer a question about collections of frames: for example, \textit{"what is the person doing?"} It computes the answer to the input question for each frame in the video and collates them. It then uses frame-wise voting to choose the most salient answer. We used voting over choosing a single frame's value in order to reduce potential for errors from a single frame. 

\noindent\textbf{\texttt{get\_summary}} Given a collection of frames, this method uses a video captioning model on discrete slices of the video and produces an overarching paragraph summary of the events in the video. We elaborate more on this capability in Section \ref{sec:long_video_summarization}. This module is useful for answering questions about the high-level narrative structure of a video, especially for longer clips such as those in Ego4D.

\noindent\textbf{\texttt{get\_script}} This computes a transcript of the audio from the input video using Whisper\cite{radford2023robust}, or returns the transcript if one already exists. This method is particularly useful for benchmarks like TVQA, where obtaining the character's dialogue is essential for solving questions.

\noindent\textbf{\texttt{get\_caption}} Computes captions for the input image or set of frames. This is useful for giving visual context of the entire scene, which can help guide choosing multiple choice answers on datasets like TVQA or NeXT-QA.

\noindent\textbf{\texttt{track\_objects}} Given a set of detections over continuous frames, associates them together using ByteTrack \cite{zhang2022bytetrack} and returns each tracked object in the scene.

\noindent\textbf{\texttt{choose\_option}} Given input context, a question and options, uses an LLM to answer a multiple choice question by choosing the most relevant answer. The input context can be visual, such as a caption or visual outputs from other modules, or textual, such as a narrative summary or transcript. This method is crucial for solving multiple choice benchmarks, and for reasoning about input context from \texttt{get\_script} or \texttt{get\_summary}.

\subsection{Long Video Summarization}
\label{sec:long_video_summarization}

One advantage of our modular approach is that we can define different modules that are better adapted to certain tasks. Consider a question that requires understanding higher-level semantics, such as \textit{"Which option best describes the overarching narrative of the video?"}. A human solving this would construct a mental narrative of the video, then match it to the list of given options. We implement a module \texttt{get\_summary()} that leverages pretrained video-to-text models to understand the high-level story of a video, which we illustrate in Figure \ref{fig:long_summarization}. Given a long video $V$ and pre-trained model $M$ that accepts a contiguous segment of frames and outputs a caption, such as LaViLa \cite{zhao2023learning}, we partition $V$ into equal-sized chunks of 1 second. For each chunk, we run $M$, outputting a caption and the timestamp of the chunk, resulting in a list of timestep annotated captions. We then aggregate this caption stream into a paragraph summary with an LLM which we use as the "narrative" of the video, with each sentence describing a 5-second interval. Importantly, this module relies on a high-quality video captioning model. This approach demonstrates strong results on the EgoSchema dataset due to the high density of annotated narrations in the underlying Ego4D dataset, enabling high-quality models like LaViLa to be trained. However, for other datasets, this approach is less effective due to the lack of comparable quality video captioning models.

\vspace{-0.5em}
\subsection{Prompting and Examples}
A long line of work \cite{wei2022chain, lewis2020retrieval} has demonstrated that the wording of the input prompt and set of examples used greatly influence LLM performance for downstream tasks  While ViperGPT includes fixed example programs in method docstrings, we follow VISPROG and CodeVQA and write a few example programs for each benchmark dataset, and modify the prompt before generation to include the most applicable examples based on the question and prompt. The in-context learning ability of LLMs results in significantly higher quality generated programs. For LLM-based modules such as \texttt{choose\_option} or \texttt{get\_summary} we keep their prompts fixed, using a static set of example inputs. All example queries and answers are sampled from training splits of benchmark datasets to avoid contamination in downstream evaluation.  We provide detailed ablations on prompt components in Section \ref{sec:ablations}, and the full prompt in Appendix \ref{sec:prompt}.

\begin{table*}
\centering
\begin{tabular}{lcccccc}
\toprule
& TGIF-QA & MSVD-QA & MSRVTT-QA & ActivityNet-QA & iVQA & TVQA \\
\midrule
Random & 0.1 & 0.1 & 0.1 & 0.1 & 0.1 & 20 \\
CLIP-VIT-L/14 \cite{radford2021learning} & 3.6 & 7.2 & 2.1 & 1.2 & 9.2 & 26.1 \\ 
Just Ask \cite{yang2021justask} & - & 13.3 & 5.6 & 12.3 & 13.3 & - \\ 
FrozenBiLM \cite{yang2022frozenbilm} & 41.9 & 33.8 & 16.9 & 25.9 & 26.8 & 59.7 \\
\textcolor{Gray}{Supervised SOTA} & \textcolor{Gray}{66.3}& \textcolor{Gray}{54.8} & \textcolor{Gray}{47.0}& \textcolor{Gray}{43.2}& \textcolor{Gray}{40.9} & \textcolor{Gray}{86.1}\\
\midrule
\methodname\. (Ours) &  \textbf{66.1(\textcolor{Green}{+25\%})} & \textbf{37.5(\textcolor{Green}{+4\%})} & \textbf{22.1(\textcolor{Green}{+5\%})} & \textbf{42.3(\textcolor{Green}{+16\%})} & \textbf{50.7(\textcolor{Green}{+23\%})} & \textbf{64.6(\textcolor{Green}{+4.9\%})}\\
\bottomrule
\end{tabular}
\caption{\textbf{Comparison of \methodname\. to state-of-the-art zero-shot video QA benchmarks.} Compared to other zero-shot methods, \methodname\. achieves state-of-the-art performance by a wide margin, improving accuracy by up to 26\% on both open-ended and multiple-choice benchmarks and is even competitive with supervised methods. }
\label{tab:zeroshot_qa_main}
\end{table*}

\begin{table}
\centering
\begin{tabular}{lc}
& Accuracy (\%)$ \uparrow$ \\
\midrule
FrozenBiLM & 26.9\\ 
VIOLET \cite{fu2021violet} & 19.9\\
mPLUG-OWL \cite{ye2023mplug} & 31.1\\
InternVideo\cite{wang2022internvideo} & 32.1\\ 
\midrule
\methodname\. (Ours) & \textbf{57.1\textcolor{Green}{(+25\%)}}
\end{tabular}
\caption{\textbf{EgoSchema results.} We report zero-shot accuracy on the held-out test split, designed to test long video understanding. \methodname\. achieves a large gain over all end-to-end models, as it can use video and language models to generate narrative summaries.}
\label{tab:egoschema_results}
\end{table}

\subsection{Open-Ended and Multiple-Choice Benchmarks}
The output answer from the program needs to be constrained to a fixed vocabulary to evaluate correctness with benchmarks. In multiple-choice QA datasets, this can be accomplished by prompting the LLM in the `\texttt{choose\_option()}` module to constrain its output to the range of input answers. On the other hand, open-ended benchmarks are typically formulated as $K$-way classification problems with $K$ ranging into the thousands. To address this, we select the semantically closest vocabulary answer to the string produced by the generated program. Concretely, we embed the output string from \texttt{video\_query} with a pre-trained phrase embedding model and find the closest match in embedding space from the output vocabulary. We used FastText \cite{joulin2017bag} for this step, and abstracted away this logic from the program execution to reduce the amount of calls required for the generated program. Furthermore, several open-ended benchmarks classify their questions into disjoint categories, such as "locations" or "objects". As the question type is available at test time, we use these splits to constrain the vocabulary for each category as well. We provide ablations over these components in Appendix \ref{sec:further_ablations}.



 \section{Experiments}


\subsection{Experimental Setup}
\label{sec:experimental_setup}
We evaluate \methodname\. on a wide variety of datasets, containing  short, long and egocentric videos with visual, narrative and multimodal questions. We briefly describe the benchmarks used here, and provide further details in Appendix \textcolor{red}{B}. Unless otherwise stated, we evaluate on the full test split of each dataset. We evaluate \methodname\. on  open-ended VideoQA (iVQA \cite{yang2021justask}, TGIF-QA FrameQA \cite{tgif-cvpr2016}, MSRVTT-QA \cite{xu2017video}, MSVD-QA \cite{xu2017video} and ActivityNet-QA \cite{yu2019activityqa}) and multiple-choice VideoQA (TVQA \cite{lei2018tvqa}, NeXT-QA \cite{xiao2021next}) and long-video understanding( EgoSchema \cite{mangalam2023egoschema}). On all datasets, we measure performance with top-1 test accuracy for fair comparison with prior work.

\begin{table}
\centering
\begin{tabular}{lc}
& Accuracy (\%)$ \uparrow$ \\
\midrule
Random & 20\\ 
ViperGPT \cite{fu2021violet} & 60.0\\
\textcolor{Gray}{Supervised SOTA} & \textcolor{Gray}{63.1}\\
\midrule
\methodname\. (Ours) & \textbf{63.8\textcolor{Green}{(+3.8\%)}}
\end{tabular}
\caption{\textbf{NeXT-QA results.} We report zero-shot accuracy on the test split. \methodname\. improves over ViperGPT's image-centric procedural method, surpassing the supervised state-of-the-art.}
\label{tab:nextqa_results}
\end{table}

\begin{figure*}[!htbp]
    \centering
    \includegraphics[width=\textwidth, height=8cm]{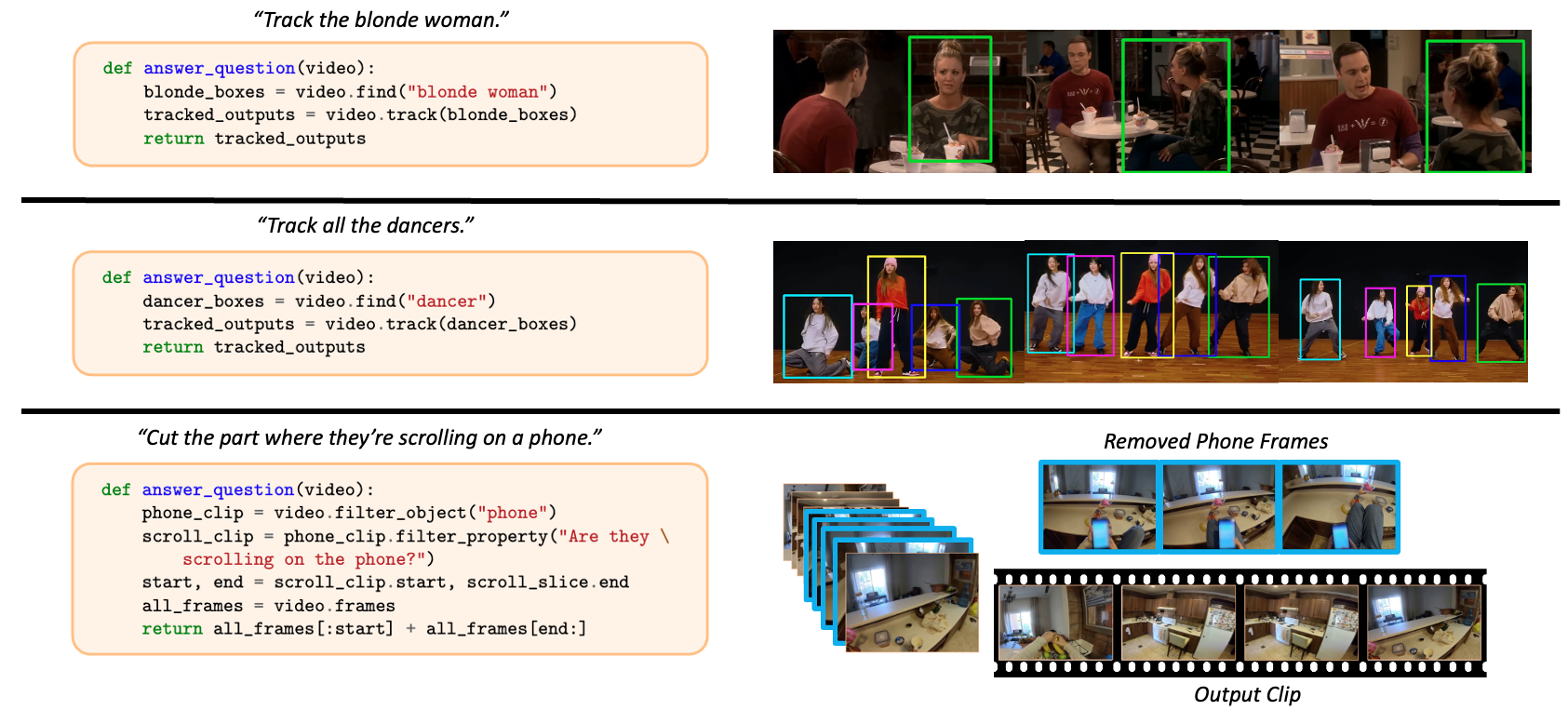}
    \caption{\textbf{\methodname\. exhibits additional video capabilities.} \methodname\. can compose its modules into programs besides question answering, such as query-based multi-object tracking or basic video editing. }
    \label{fig:vidprog_capabilities}
\end{figure*}

\begin{table*}
\centering
\begin{tabular}{ccc@{\,\,}c@{\,\,}c@{\,\,}c|c@{\,\,}c@{\,\,}c@{\,\,}c@{\,\,}c@{\,\,}c}
\toprule
\texttt{IQ} & \texttt{VQ} & \texttt{FP} & \texttt{find} & \texttt{caption} & \texttt{full} & TGIF-QA & MSVD-QA & MSRVTT-QA & ActivityNet-QA & iVQA  & NeXT-QA\\
\midrule
\checkmark & & & & & & 62.1 & 34.1 & 16.6  & 27.7 & 41.5 & 49.3\\
&  \checkmark& & & & & 63.4 & \textbf{37.8} & 20.1 & 35.0 & 46.5 & 53.2 \\ 
&  \checkmark&  \checkmark&  & & & 66.1 & 37.5 & 24.1 & 39.1 & 50.3 & 55.2 \\ 
&  \checkmark&  \checkmark&  \checkmark&  & & 66.1 & 37.5 & \textbf{23.5} & \textbf{42.6} & \textbf{52.0} & 55.9 \\
&  \checkmark&  \checkmark&  \checkmark&  \checkmark& & 66.1 & 22.8 & 16.9 & 42.3 & 50.7 & 63.8 \\ 
\midrule
 &  \checkmark&  \checkmark&  \checkmark&  \checkmark&  \checkmark&  \textbf{66.1} & 37.5 & 22.1 & 42.3 & 50.7 & \textbf{64.6} \\
\bottomrule
\end{tabular}
\caption{\textbf{Ablating the visual modules.}  We successively ablate the performance with single-image querying only (\texttt{IQ}), video querying (\texttt{VQ}), filtering (\texttt{FP}), the \texttt{find}, \texttt{caption} modules, and then the full prompt. Generally, adding more modules can slightly reduce performance on individual datasets, but enables generalization to a much wider range of benchmarks.}
\label{tab:module_ablation}
\end{table*}

\subsection{Zero-Shot Video QA Results}
\label{sec:zeroshot_vqa}
Our main results are contained in Table \ref{tab:zeroshot_qa_main} and Table \ref{tab:egoschema_results}. We achieve large accuracy improvements on both open-ended and multiple choice benchmarks, with particularly large gains of 23\% on iVQA, 25\% on TGIF-QA and 25\% on the challenging EgoSchema benchmark. We attribute these improvements to a few different factors. 
On open-ended benchmarks, such as TGIF, ActivityNet, and iVQA, we observe much larger improvements as \methodname\. can effectively find the relevant video segments through its generated program, then use strong vision-language models only on n that segment to compute the answer. In contrast, end-to-end methods like Just Ask and FrozenBiLM consider all frames and are less able to focus on the informative segments of the video. We also outperform ViperGPT, which uses an image-centric API and is unable to reason over longer intervals of time.
We also observe that performance improvements were highly correlated with dataset label quality. MSR-VTT and  MSVD have a large fraction of ambiguously or incorrectly labeled questions, while iVQA, TGIF-QA and ActivityNet-QA have higher quality labels. In particular, iVQA contains multiple correct answers per question, making the vocabulary-matching much more forgiving. On multiple-choice datasets, such as TVQA and NeXT-QA, the improvement is less due to the constraints imposed by having much fewer choices; on TVQA in particular, most questions are dominated by the language model reasoning over the input script rather than visual elements, leading to a smaller but still significant improvement. 

\subsection{Additional Capabilities}
\methodname\. can use its modules to perform other video tasks by composing its modules in different ways. We provide qualitative examples in Figure \ref{fig:vidprog_capabilities} with \methodname\.'s ability and perform multi-object tracking or video editing. Other tasks are straightforward to implement as well, requiring only a module for the specific functionality.

\label{sec:tracking_editing}
\noindent\textbf{Grounded Tracking} Since \methodname\. can use a text-conditioned object detector, it can combine it with the tracking module to track multiple objects in a scene based on an input query. An example is shown in Figure \ref{fig:vidprog_samples} where we are able to track all the dancers in a complicated scene through detecting them in each frame, and tracking them over time. 

\noindent\textbf{Video Editing} \methodname\. can also combine its modules to retrieve relevant clips and remove clips that do not satisfy input criteria, enabling a basic form of video editing. An example is shown in Figure \ref{fig:vidprog_capabilities}, where we ask \methodname\. to cut all parts of a video where the subject is scrolling their phone.

\subsection{Ablation Studies}
\label{sec:ablations}

\begin{figure}
    \centering
    \includegraphics[width=\linewidth]{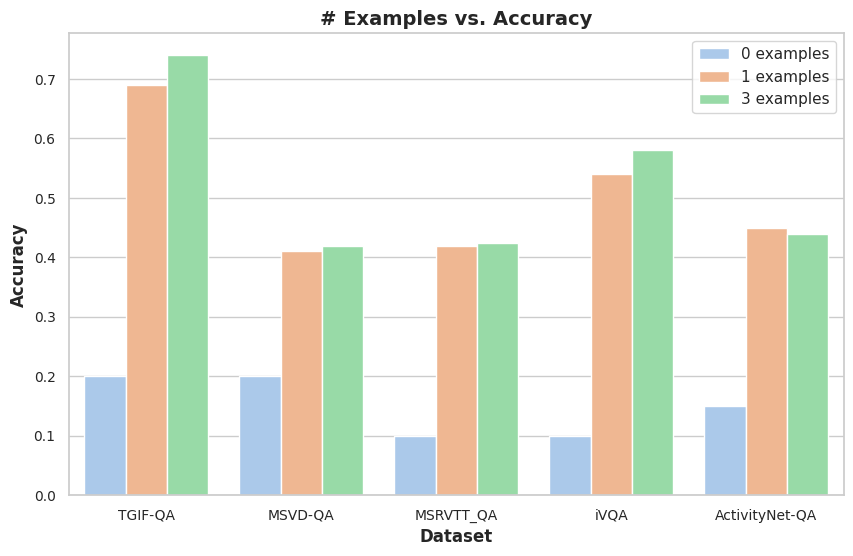}
    \caption{\textbf{Impact of in-context examples.} Without in-context example programs, \methodname\. performs poorly, but adding a single example increases performance significantly. More examples further improve accuracy, but with diminishing returns. }
    \label{fig:ablate_examples}
\end{figure}

\noindent\textbf{Individual Modules.} We ablate each module to better understand the observed boosts in performance, shown shown in Table \ref{tab:module_ablation}. 
We omit the \texttt{get\_summary} module from this analysis since it only applies to the EgoSchema benchmark; the other datasets do not have sufficiently performant video-to-text models.
Including a image QA module already provides a strong baseline, suggesting that simply prompting a strong image QA model with the right questions can be helpful. Including the \texttt{video\_query} module, enabling QA over video segments, results in a large boost as it is less prone to errors from querying a single frame; this accounts for most of the improvement in shorter video datasets. The \texttt{filter\_property} and \texttt{filter\_object} modules especially help on longer video datasets, such as ActivityNet and NeXT-QA. The \texttt{find} module helps answer ``counting" and ``color" questions in ActivityNet, explaining the observed accuracy increase, and the \texttt{get\_caption} module adds a significant boost on NeXT-QA, which needs visual descriptions of scenes to answer causal questions. In general, adding more and more modules slightly decreases performance, as the language model can misuse modules or combine them in ways that may not compile. However, this effect is mitigated with in-context examples, and including more modules allows \methodname\, to generalize to a wider variety of benchmarks with state-of-the-art performance.

\noindent\textbf{In-context Examples.} We next ablate the in-context examples provided in the prompt to understand their effect. In Figure \ref{fig:ablate_examples}, we measure the effect of varying the number of in-context examples on downstream benchmarks. With no examples, performance is poor: the LLM generates programs that may not compile, hallucinates methods or writes overly complex code. Adding a single example program greatly improves performance, and while including more examples is helpful, it provides diminishing returns. We found that using 3-4 examples worked well across benchmarks. 

\begin{figure}
    \centering
    \includegraphics[width=0.8\linewidth]{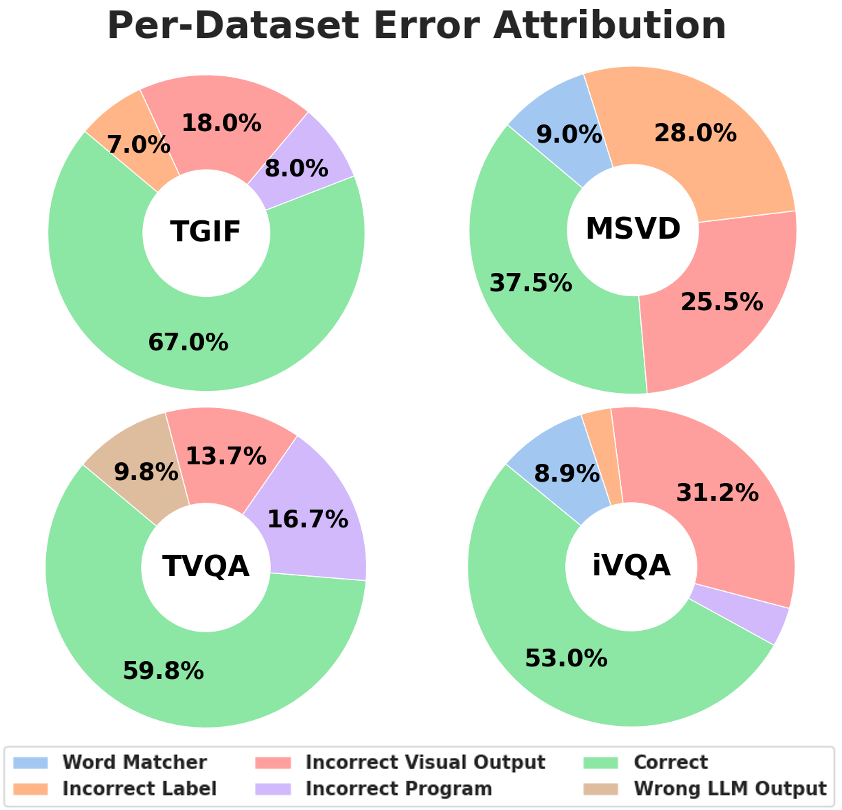}
    \caption{\textbf{Error Analysis.} On lower-quality or open-ended datasets, \methodname\. suffers mostly from visual module failure and labeling issues. On multiple-choice settings, the generated program is usually the issue, while in higher-quality open-ended benchmarks, the visual modules are more reliable. }
    \label{fig:error_analysis}
\end{figure}

\noindent\textbf{Failure Modes.} We manually inspected the failure modes of \methodname\., shown in Figure \ref{fig:error_analysis}, on 100 random samples from each dataset. Due to the interpretability of our method, we can effectively attribute the cause of errors to either an incorrect program, module failure, or incorrect labeling. We conduct this analysis on MSR-VTT, ActivityNet, iVQA and TVQA. On open-ended benchmarks, a considerable portion of the dataset are ambiguously or incorrectly labeled, with \methodname\. outputting a correct answer. We find that the balance of program generation vs. module failures depends on the dataset: lower quality and shorter video datasets, such as MSVD and MSR-VTT, mostly suffer from annotation error or mistakes from individual visual modules. 
Although iVQA has much higher annotation quality, the word-matching embedding model often misclassifies the output from \texttt{video\_query}, leading to correct output from the program but mistakes in post-processing. 
On the other hand, \methodname\.'s mistakes on TVQA, a multimodal dataset, are mostly  at the program generation phase, as the language model often uses the the wrong modalities, such as checking the speech transcript to answer visual questions.

 \section{Conclusions}
We presented \methodname\., a method that extends modular vision methods to zero-shot video question answering. Our approach achieves a significant improvement over the state-of-the-art by using strong language models for temporal reasoning with code rather than using an end-to-end network. Extensive ablation studies and analysis demonstrate the strengths of our method, suggesting procedural reasoning can improve performance for vision tasks in general. We also present other creative capabilities enabled by \methodname\. such as grounded tracking and text-based video editing. that require no modifications to the method. We believe \methodname\. will serve as a strong video question answering baseline for the community going forward.

\noindent\textbf{Acknowledgements.} This work was partially supported by Fujitsu Research of America. RC is supported by the NSF GRFP.

 {
     \small
     \bibliographystyle{ieeenat_fullname}
     \bibliography{main}

\begin{thebibliography}{63}
\providecommand{\natexlab}[1]{#1}
\providecommand{\url}[1]{\texttt{#1}}
\expandafter\ifx\csname urlstyle\endcsname\relax
  \providecommand{\doi}[1]{doi: #1}\else
  \providecommand{\doi}{doi: \begingroup \urlstyle{rm}\Url}\fi

\bibitem[Alayrac et~al.(2022)Alayrac, Donahue, Luc, Miech, Barr, Hasson, Lenc, Mensch, Millican, Reynolds, et~al.]{alayrac2022flamingo}
Jean-Baptiste Alayrac, Jeff Donahue, Pauline Luc, Antoine Miech, Iain Barr, Yana Hasson, Karel Lenc, Arthur Mensch, Katherine Millican, Malcolm Reynolds, et~al.
\newblock Flamingo: a visual language model for few-shot learning.
\newblock \emph{Advances in Neural Information Processing Systems}, 35:\penalty0 23716--23736, 2022.

\bibitem[Andreas et~al.(2016)Andreas, Rohrbach, Darrell, and Klein]{andreas2016neural}
Jacob Andreas, Marcus Rohrbach, Trevor Darrell, and Dan Klein.
\newblock Neural module networks.
\newblock In \emph{CVPR}, 2016.

\bibitem[Bain et~al.(2021)Bain, Nagrani, Varol, and Zisserman]{bain2021frozen}
Max Bain, Arsha Nagrani, G{\"u}l Varol, and Andrew Zisserman.
\newblock Frozen in time: A joint video and image encoder for end-to-end retrieval.
\newblock In \emph{Proceedings of the IEEE/CVF International Conference on Computer Vision}, pages 1728--1738, 2021.

\bibitem[Brown et~al.(2020)Brown, Mann, Ryder, Subbiah, Kaplan, Dhariwal, Neelakantan, Shyam, Sastry, Askell, et~al.]{brown2020language}
Tom Brown, Benjamin Mann, Nick Ryder, Melanie Subbiah, Jared~D Kaplan, Prafulla Dhariwal, Arvind Neelakantan, Pranav Shyam, Girish Sastry, Amanda Askell, et~al.
\newblock Language models are few-shot learners.
\newblock \emph{Advances in neural information processing systems}, 33:\penalty0 1877--1901, 2020.

\bibitem[Carreira and Zisserman(2017)]{carreira2017quo}
Joao Carreira and Andrew Zisserman.
\newblock Quo vadis, action recognition? a new model and the kinetics dataset.
\newblock In \emph{proceedings of the IEEE Conference on Computer Vision and Pattern Recognition}, pages 6299--6308, 2017.

\bibitem[Devlin et~al.(2018)Devlin, Chang, Lee, and Toutanova]{devlin2018bert}
Jacob Devlin, Ming-Wei Chang, Kenton Lee, and Kristina Toutanova.
\newblock Bert: Pre-training of deep bidirectional transformers for language understanding.
\newblock \emph{arXiv preprint arXiv:1810.04805}, 2018.

\bibitem[Fan et~al.(2019)Fan, Zhang, Zhang, Wang, Zhang, and Huang]{fan2019heterogeneous}
Chenyou Fan, Xiaofan Zhang, Shu Zhang, Wensheng Wang, Chi Zhang, and Heng Huang.
\newblock Heterogeneous memory enhanced multimodal attention model for video question answering.
\newblock In \emph{Proceedings of the IEEE/CVF conference on computer vision and pattern recognition}, pages 1999--2007, 2019.

\bibitem[Fu et~al.(2021)Fu, Li, Gan, Lin, Wang, Wang, and Liu]{fu2021violet}
Tsu-Jui Fu, Linjie Li, Zhe Gan, Kevin Lin, William~Yang Wang, Lijuan Wang, and Zicheng Liu.
\newblock Violet: End-to-end video-language transformers with masked visual-token modeling.
\newblock \emph{arXiv preprint arXiv:2111.12681}, 2021.

\bibitem[Gao et~al.(2018)Gao, Ge, Chen, and Nevatia]{gao2018motion}
Jiyang Gao, Runzhou Ge, Kan Chen, and Ram Nevatia.
\newblock Motion-appearance co-memory networks for video question answering.
\newblock In \emph{Proceedings of the IEEE Conference on Computer Vision and Pattern Recognition}, pages 6576--6585, 2018.

\bibitem[Girdhar et~al.(2023)Girdhar, El-Nouby, Liu, Singh, Alwala, Joulin, and Misra]{girdhar2023imagebind}
Rohit Girdhar, Alaaeldin El-Nouby, Zhuang Liu, Mannat Singh, Kalyan~Vasudev Alwala, Armand Joulin, and Ishan Misra.
\newblock Imagebind: One embedding space to bind them all.
\newblock In \emph{CVPR}, 2023.

\bibitem[Grauman et~al.(2022)Grauman, Westbury, Byrne, Chavis, Furnari, Girdhar, Hamburger, Jiang, Liu, Liu, et~al.]{grauman2022ego4d}
Kristen Grauman, Andrew Westbury, Eugene Byrne, Zachary Chavis, Antonino Furnari, Rohit Girdhar, Jackson Hamburger, Hao Jiang, Miao Liu, Xingyu Liu, et~al.
\newblock Ego4d: Around the world in 3,000 hours of egocentric video.
\newblock In \emph{Proceedings of the IEEE/CVF Conference on Computer Vision and Pattern Recognition}, pages 18995--19012, 2022.

\bibitem[Gupta and Kembhavi(2023)]{gupta2023visual}
Tanmay Gupta and Aniruddha Kembhavi.
\newblock Visual programming: Compositional visual reasoning without training.
\newblock In \emph{CVPR}, 2023.

\bibitem[Hu et~al.(2017)Hu, Andreas, Rohrbach, Darrell, and Saenko]{hu2017learning}
Ronghang Hu, Jacob Andreas, Marcus Rohrbach, Trevor Darrell, and Kate Saenko.
\newblock Learning to reason: End-to-end module networks for visual question answering.
\newblock In \emph{ICCV}, 2017.

\bibitem[Hu et~al.(2018)Hu, Andreas, Darrell, and Saenko]{hu2018explainable}
Ronghang Hu, Jacob Andreas, Trevor Darrell, and Kate Saenko.
\newblock Explainable neural computation via stack neural module networks.
\newblock In \emph{Proceedings of the European conference on computer vision (ECCV)}, pages 53--69, 2018.

\bibitem[Huang et~al.(2020)Huang, Chen, Zeng, Du, Tan, and Gan]{huang2020location}
Deng Huang, Peihao Chen, Runhao Zeng, Qing Du, Mingkui Tan, and Chuang Gan.
\newblock Location-aware graph convolutional networks for video question answering.
\newblock In \emph{Proceedings of the AAAI Conference on Artificial Intelligence}, pages 11021--11028, 2020.

\bibitem[Jang et~al.(2017)Jang, Song, Yu, Kim, and Kim]{jang2017tgif}
Yunseok Jang, Yale Song, Youngjae Yu, Youngjin Kim, and Gunhee Kim.
\newblock Tgif-qa: Toward spatio-temporal reasoning in visual question answering.
\newblock In \emph{Proceedings of the IEEE conference on computer vision and pattern recognition}, pages 2758--2766, 2017.

\bibitem[Joulin et~al.(2017)Joulin, Grave, Bojanowski, and Mikolov]{joulin2017bag}
Armand Joulin, Edouard Grave, Piotr Bojanowski, and Tomas Mikolov.
\newblock Bag of tricks for efficient text classification.
\newblock In \emph{Proceedings of the 15th Conference of the European Chapter of the Association for Computational Linguistics: Volume 2, Short Papers}, pages 427--431. Association for Computational Linguistics, 2017.

\bibitem[Kim et~al.(2020)Kim, Ma, Pham, Kim, and Yoo]{kim2020modality}
Junyeong Kim, Minuk Ma, Trung Pham, Kyungsu Kim, and Chang~D Yoo.
\newblock Modality shifting attention network for multi-modal video question answering.
\newblock In \emph{Proceedings of the IEEE/CVF conference on computer vision and pattern recognition}, pages 10106--10115, 2020.

\bibitem[Le et~al.(2020)Le, Le, Venkatesh, and Tran]{le2020hierarchical}
Thao~Minh Le, Vuong Le, Svetha Venkatesh, and Truyen Tran.
\newblock Hierarchical conditional relation networks for video question answering.
\newblock In \emph{Proceedings of the IEEE/CVF conference on computer vision and pattern recognition}, pages 9972--9981, 2020.

\bibitem[Lei et~al.(2018)Lei, Yu, Bansal, and Berg]{lei2018tvqa}
Jie Lei, Licheng Yu, Mohit Bansal, and Tamara~L Berg.
\newblock Tvqa: Localized, compositional video question answering.
\newblock \emph{arXiv preprint arXiv:1809.01696}, 2018.

\bibitem[Lewis et~al.(2020)Lewis, Perez, Piktus, Petroni, Karpukhin, Goyal, K{\"u}ttler, Lewis, Yih, Rockt{\"a}schel, et~al.]{lewis2020retrieval}
Patrick Lewis, Ethan Perez, Aleksandra Piktus, Fabio Petroni, Vladimir Karpukhin, Naman Goyal, Heinrich K{\"u}ttler, Mike Lewis, Wen-tau Yih, Tim Rockt{\"a}schel, et~al.
\newblock Retrieval-augmented generation for knowledge-intensive nlp tasks.
\newblock \emph{Advances in Neural Information Processing Systems}, 33:\penalty0 9459--9474, 2020.

\bibitem[Li et~al.(2022)Li, Li, Xiong, and Hoi]{li2022blip}
Junnan Li, Dongxu Li, Caiming Xiong, and Steven Hoi.
\newblock Blip: Bootstrapping language-image pre-training for unified vision-language understanding and generation.
\newblock In \emph{International Conference on Machine Learning}, pages 12888--12900. PMLR, 2022.

\bibitem[Li et~al.(2023{\natexlab{a}})Li, Li, Savarese, and Hoi]{li2023blip}
Junnan Li, Dongxu Li, Silvio Savarese, and Steven Hoi.
\newblock Blip-2: Bootstrapping language-image pre-training with frozen image encoders and large language models.
\newblock \emph{arXiv preprint arXiv:2301.12597}, 2023{\natexlab{a}}.

\bibitem[Li et~al.(2023{\natexlab{b}})Li, He, Wang, Li, Wang, Luo, Wang, Wang, and Qiao]{2023videochat}
Kunchang Li, Yinan He, Yi Wang, Yizhuo Li, Wenhai Wang, Ping Luo, Yali Wang, Limin Wang, and Yu Qiao.
\newblock Videochat: Chat-centric video understanding.
\newblock \emph{arXiv preprint arXiv:2305.06355}, 2023{\natexlab{b}}.

\bibitem[Li et~al.(2016)Li, Song, Cao, Tetreault, Goldberg, Jaimes, and Luo]{tgif-cvpr2016}
Yuncheng Li, Yale Song, Liangliang Cao, Joel Tetreault, Larry Goldberg, Alejandro Jaimes, and Jiebo Luo.
\newblock {TGIF: A New Dataset and Benchmark on Animated GIF Description}.
\newblock In \emph{The IEEE Conference on Computer Vision and Pattern Recognition (CVPR)}, 2016.

\bibitem[Lin et~al.(2021)Lin, Bertasius, Wang, Chang, Parikh, and Torresani]{lin2021vx2text}
Xudong Lin, Gedas Bertasius, Jue Wang, Shih-Fu Chang, Devi Parikh, and Lorenzo Torresani.
\newblock Vx2text: End-to-end learning of video-based text generation from multimodal inputs.
\newblock In \emph{Proceedings of the IEEE/CVF Conference on Computer Vision and Pattern Recognition}, pages 7005--7015, 2021.

\bibitem[Liu et~al.(2023{\natexlab{a}})Liu, Li, Li, and Lee]{liu2023improvedllava}
Haotian Liu, Chunyuan Li, Yuheng Li, and Yong~Jae Lee.
\newblock Improved baselines with visual instruction tuning, 2023{\natexlab{a}}.

\bibitem[Liu et~al.(2023{\natexlab{b}})Liu, Li, Wu, and Lee]{liu2023llava}
Haotian Liu, Chunyuan Li, Qingyang Wu, and Yong~Jae Lee.
\newblock Visual instruction tuning.
\newblock In \emph{NeurIPS}, 2023{\natexlab{b}}.

\bibitem[Liu et~al.(2023{\natexlab{c}})Liu, Zeng, Ren, Li, Zhang, Yang, Li, Yang, Su, Zhu, et~al.]{liu2023grounding}
Shilong Liu, Zhaoyang Zeng, Tianhe Ren, Feng Li, Hao Zhang, Jie Yang, Chunyuan Li, Jianwei Yang, Hang Su, Jun Zhu, et~al.
\newblock Grounding dino: Marrying dino with grounded pre-training for open-set object detection.
\newblock \emph{arXiv preprint arXiv:2303.05499}, 2023{\natexlab{c}}.

\bibitem[Maaz et~al.(2023)Maaz, Rasheed, Khan, and Khan]{Maaz2023VideoChatGPT}
Muhammad Maaz, Hanoona Rasheed, Salman Khan, and Fahad Khan.
\newblock Video-chatgpt: Towards detailed video understanding via large vision and language models.
\newblock \emph{ArXiv 2306.05424}, 2023.

\bibitem[Mangalam et~al.(2023)Mangalam, Akshulakov, and Malik]{mangalam2023egoschema}
Karttikeya Mangalam, Raiymbek Akshulakov, and Jitendra Malik.
\newblock Egoschema: A diagnostic benchmark for very long-form video language understanding.
\newblock \emph{arXiv preprint arXiv:2308.09126}, 2023.

\bibitem[Miech et~al.(2019)Miech, Zhukov, Alayrac, Tapaswi, Laptev, and Sivic]{miech2019howto100m}
Antoine Miech, Dimitri Zhukov, Jean-Baptiste Alayrac, Makarand Tapaswi, Ivan Laptev, and Josef Sivic.
\newblock Howto100m: Learning a text-video embedding by watching hundred million narrated video clips.
\newblock In \emph{Proceedings of the IEEE/CVF international conference on computer vision}, pages 2630--2640, 2019.

\bibitem[Mikolov et~al.(2013)Mikolov, Chen, Corrado, and Dean]{mikolov2013efficient}
Tomas Mikolov, Kai Chen, Greg Corrado, and Jeffrey Dean.
\newblock Efficient estimation of word representations in vector space.
\newblock \emph{arXiv preprint arXiv:1301.3781}, 2013.

\bibitem[OpenAI(2023)]{openai2023gpt}
R OpenAI.
\newblock Gpt-4 technical report.
\newblock \emph{arXiv}, pages 2303--08774, 2023.

\bibitem[Park et~al.(2021)Park, Lee, and Sohn]{park2021bridge}
Jungin Park, Jiyoung Lee, and Kwanghoon Sohn.
\newblock Bridge to answer: Structure-aware graph interaction network for video question answering.
\newblock In \emph{Proceedings of the IEEE/CVF conference on computer vision and pattern recognition}, pages 15526--15535, 2021.

\bibitem[Radford et~al.(2021)Radford, Kim, Hallacy, Ramesh, Goh, Agarwal, Sastry, Askell, Mishkin, Clark, et~al.]{radford2021learning}
Alec Radford, Jong~Wook Kim, Chris Hallacy, Aditya Ramesh, Gabriel Goh, Sandhini Agarwal, Girish Sastry, Amanda Askell, Pamela Mishkin, Jack Clark, et~al.
\newblock Learning transferable visual models from natural language supervision.
\newblock In \emph{International conference on machine learning}, pages 8748--8763. PMLR, 2021.

\bibitem[Radford et~al.(2023)Radford, Kim, Xu, Brockman, McLeavey, and Sutskever]{radford2023robust}
Alec Radford, Jong~Wook Kim, Tao Xu, Greg Brockman, Christine McLeavey, and Ilya Sutskever.
\newblock Robust speech recognition via large-scale weak supervision.
\newblock In \emph{International Conference on Machine Learning}, pages 28492--28518. PMLR, 2023.

\bibitem[Reddy et~al.(2022)Reddy, Rui, Li, Lin, Wen, Cho, Huang, Bansal, Sil, Chang, et~al.]{reddy2022mumuqa}
Revant~Gangi Reddy, Xilin Rui, Manling Li, Xudong Lin, Haoyang Wen, Jaemin Cho, Lifu Huang, Mohit Bansal, Avirup Sil, Shih-Fu Chang, et~al.
\newblock Mumuqa: Multimedia multi-hop news question answering via cross-media knowledge extraction and grounding.
\newblock In \emph{Proceedings of the AAAI Conference on Artificial Intelligence}, pages 11200--11208, 2022.

\bibitem[Saqur and Narasimhan(2020)]{saqur2020multimodal}
Raeid Saqur and Karthik Narasimhan.
\newblock Multimodal graph networks for compositional generalization in visual question answering.
\newblock \emph{Advances in Neural Information Processing Systems}, 33:\penalty0 3070--3081, 2020.

\bibitem[Schick et~al.(2023)Schick, Dwivedi-Yu, Dess{\`\i}, Raileanu, Lomeli, Zettlemoyer, Cancedda, and Scialom]{schick2023toolformer}
Timo Schick, Jane Dwivedi-Yu, Roberto Dess{\`\i}, Roberta Raileanu, Maria Lomeli, Luke Zettlemoyer, Nicola Cancedda, and Thomas Scialom.
\newblock Toolformer: Language models can teach themselves to use tools.
\newblock \emph{arXiv preprint arXiv:2302.04761}, 2023.

\bibitem[Seo et~al.(2021)Seo, Nagrani, and Schmid]{seo2021look}
Paul~Hongsuck Seo, Arsha Nagrani, and Cordelia Schmid.
\newblock Look before you speak: Visually contextualized utterances.
\newblock In \emph{Proceedings of the IEEE/CVF Conference on Computer Vision and Pattern Recognition}, pages 16877--16887, 2021.

\bibitem[Song et~al.(2023)Song, Chai, Wang, Zhang, Zhou, Wu, Guo, Ye, Lu, Hwang, et~al.]{song2023moviechat}
Enxin Song, Wenhao Chai, Guanhong Wang, Yucheng Zhang, Haoyang Zhou, Feiyang Wu, Xun Guo, Tian Ye, Yan Lu, Jenq-Neng Hwang, et~al.
\newblock Moviechat: From dense token to sparse memory for long video understanding.
\newblock \emph{arXiv preprint arXiv:2307.16449}, 2023.

\bibitem[Subramanian et~al.(2023)Subramanian, Narasimhan, Khangaonkar, Yang, Nagrani, Schmid, Zeng, Darrell, and Klein]{subramanian-etal-2023-modular}
Sanjay Subramanian, Medhini Narasimhan, Kushal Khangaonkar, Kevin Yang, Arsha Nagrani, Cordelia Schmid, Andy Zeng, Trevor Darrell, and Dan Klein.
\newblock Modular visual question answering via code generation.
\newblock In \emph{Proceedings of the 61st Annual Meeting of the Association for Computational Linguistics (Volume 2: Short Papers)}, pages 747--761, Toronto, Canada, 2023. Association for Computational Linguistics.

\bibitem[Sur{\'\i}s et~al.(2023)Sur{\'\i}s, Menon, and Vondrick]{Suris_2023_ICCV}
D{\'\i}dac Sur{\'\i}s, Sachit Menon, and Carl Vondrick.
\newblock Vipergpt: Visual inference via python execution for reasoning.
\newblock In \emph{Proceedings of the IEEE/CVF International Conference on Computer Vision (ICCV)}, pages 11888--11898, 2023.

\bibitem[Wang et~al.(2022{\natexlab{a}})Wang, Li, Li, He, Huang, Zhao, Zhang, Xu, Liu, Wang, et~al.]{wang2022internvideo}
Yi Wang, Kunchang Li, Yizhuo Li, Yinan He, Bingkun Huang, Zhiyu Zhao, Hongjie Zhang, Jilan Xu, Yi Liu, Zun Wang, et~al.
\newblock Internvideo: General video foundation models via generative and discriminative learning.
\newblock \emph{arXiv preprint arXiv:2212.03191}, 2022{\natexlab{a}}.

\bibitem[Wang et~al.(2022{\natexlab{b}})Wang, Li, Xu, Zhou, Lei, Lin, Wang, Yang, Zhu, Hoiem, et~al.]{wang2022language}
Zhenhailong Wang, Manling Li, Ruochen Xu, Luowei Zhou, Jie Lei, Xudong Lin, Shuohang Wang, Ziyi Yang, Chenguang Zhu, Derek Hoiem, et~al.
\newblock Language models with image descriptors are strong few-shot video-language learners.
\newblock \emph{Advances in Neural Information Processing Systems}, 35:\penalty0 8483--8497, 2022{\natexlab{b}}.

\bibitem[Wei et~al.(2022)Wei, Wang, Schuurmans, Bosma, Xia, Chi, Le, Zhou, et~al.]{wei2022chain}
Jason Wei, Xuezhi Wang, Dale Schuurmans, Maarten Bosma, Fei Xia, Ed Chi, Quoc~V Le, Denny Zhou, et~al.
\newblock Chain-of-thought prompting elicits reasoning in large language models.
\newblock \emph{Advances in Neural Information Processing Systems}, 35:\penalty0 24824--24837, 2022.

\bibitem[Xiao et~al.(2021)Xiao, Shang, Yao, and Chua]{xiao2021next}
Junbin Xiao, Xindi Shang, Angela Yao, and Tat-Seng Chua.
\newblock Next-qa: Next phase of question-answering to explaining temporal actions.
\newblock In \emph{Proceedings of the IEEE/CVF conference on computer vision and pattern recognition}, pages 9777--9786, 2021.

\bibitem[Xu et~al.(2017)Xu, Zhao, Xiao, Wu, Zhang, He, and Zhuang]{xu2017video}
Dejing Xu, Zhou Zhao, Jun Xiao, Fei Wu, Hanwang Zhang, Xiangnan He, and Yueting Zhuang.
\newblock Video question answering via gradually refined attention over appearance and motion.
\newblock In \emph{ACM Multimedia}, 2017.

\bibitem[Yang et~al.(2021)Yang, Miech, Sivic, Laptev, and Schmid]{yang2021justask}
Antoine Yang, Antoine Miech, Josef Sivic, Ivan Laptev, and Cordelia Schmid.
\newblock Just ask: Learning to answer questions from millions of narrated videos.
\newblock In \emph{ICCV}, 2021.

\bibitem[Yang et~al.(2022{\natexlab{a}})Yang, Miech, Sivic, Laptev, and Schmid]{yang2022frozenbilm}
Antoine Yang, Antoine Miech, Josef Sivic, Ivan Laptev, and Cordelia Schmid.
\newblock Zero-shot video question answering via frozen bidirectional language models.
\newblock In \emph{NeurIPS}, 2022{\natexlab{a}}.

\bibitem[Yang et~al.(2023)Yang, Song, Li, Zhao, Ge, Li, and Shan]{yang2023gpt4tools}
Rui Yang, Lin Song, Yanwei Li, Sijie Zhao, Yixiao Ge, Xiu Li, and Ying Shan.
\newblock Gpt4tools: Teaching large language model to use tools via self-instruction.
\newblock \emph{arXiv preprint arXiv:2305.18752}, 2023.

\bibitem[Yang et~al.(2022{\natexlab{b}})Yang, Gan, Wang, Hu, Lu, Liu, and Wang]{yang2021empirical}
Zhengyuan Yang, Zhe Gan, Jianfeng Wang, Xiaowei Hu, Yumao Lu, Zicheng Liu, and Lijuan Wang.
\newblock An empirical study of gpt-3 for few-shot knowledge-based vqa.
\newblock In \emph{AAAI}, 2022{\natexlab{b}}.

\bibitem[Ye et~al.(2023{\natexlab{a}})Ye, Xu, Yan, Xu, Qian, Zhang, and Huang]{ye2023hitea}
Qinghao Ye, Guohai Xu, Ming Yan, Haiyang Xu, Qi Qian, Ji Zhang, and Fei Huang.
\newblock Hitea: Hierarchical temporal-aware video-language pre-training.
\newblock In \emph{Proceedings of the IEEE/CVF International Conference on Computer Vision}, pages 15405--15416, 2023{\natexlab{a}}.

\bibitem[Ye et~al.(2023{\natexlab{b}})Ye, Xu, Xu, Ye, Yan, Zhou, Wang, Hu, Shi, Shi, et~al.]{ye2023mplug}
Qinghao Ye, Haiyang Xu, Guohai Xu, Jiabo Ye, Ming Yan, Yiyang Zhou, Junyang Wang, Anwen Hu, Pengcheng Shi, Yaya Shi, et~al.
\newblock mplug-owl: Modularization empowers large language models with multimodality.
\newblock \emph{arXiv preprint arXiv:2304.14178}, 2023{\natexlab{b}}.

\bibitem[Yu et~al.(2019)Yu, Xu, Yu, Yu, Zhao, Zhuang, and Tao]{yu2019activityqa}
Zhou Yu, Dejing Xu, Jun Yu, Ting Yu, Zhou Zhao, Yueting Zhuang, and Dacheng Tao.
\newblock Activitynet-qa: A dataset for understanding complex web videos via question answering.
\newblock In \emph{AAAI}, pages 9127--9134, 2019.

\bibitem[Zellers et~al.(2021)Zellers, Lu, Hessel, Yu, Park, Cao, Farhadi, and Choi]{zellers2021merlot}
Rowan Zellers, Ximing Lu, Jack Hessel, Youngjae Yu, Jae~Sung Park, Jize Cao, Ali Farhadi, and Yejin Choi.
\newblock Merlot: Multimodal neural script knowledge models.
\newblock \emph{Advances in Neural Information Processing Systems}, 34:\penalty0 23634--23651, 2021.

\bibitem[Zellers et~al.(2022)Zellers, Lu, Lu, Yu, Zhao, Salehi, Kusupati, Hessel, Farhadi, and Choi]{zellers2022merlot}
Rowan Zellers, Jiasen Lu, Ximing Lu, Youngjae Yu, Yanpeng Zhao, Mohammadreza Salehi, Aditya Kusupati, Jack Hessel, Ali Farhadi, and Yejin Choi.
\newblock Merlot reserve: Neural script knowledge through vision and language and sound.
\newblock In \emph{Proceedings of the IEEE/CVF Conference on Computer Vision and Pattern Recognition}, pages 16375--16387, 2022.

\bibitem[Zeng et~al.(2022)Zeng, Attarian, Ichter, Choromanski, Wong, Welker, Tombari, Purohit, Ryoo, Sindhwani, et~al.]{zeng2022socratic}
Andy Zeng, Maria Attarian, Brian Ichter, Krzysztof Choromanski, Adrian Wong, Stefan Welker, Federico Tombari, Aveek Purohit, Michael Ryoo, Vikas Sindhwani, et~al.
\newblock Socratic models: Composing zero-shot multimodal reasoning with language.
\newblock \emph{arXiv preprint arXiv:2204.00598}, 2022.

\bibitem[Zhang et~al.(2022)Zhang, Sun, Jiang, Yu, Weng, Yuan, Luo, Liu, and Wang]{zhang2022bytetrack}
Yifu Zhang, Peize Sun, Yi Jiang, Dongdong Yu, Fucheng Weng, Zehuan Yuan, Ping Luo, Wenyu Liu, and Xinggang Wang.
\newblock Bytetrack: Multi-object tracking by associating every detection box.
\newblock 2022.

\bibitem[Zhao et~al.(2023)Zhao, Misra, Kr{\"a}henb{\"u}hl, and Girdhar]{zhao2023learning}
Yue Zhao, Ishan Misra, Philipp Kr{\"a}henb{\"u}hl, and Rohit Girdhar.
\newblock Learning video representations from large language models.
\newblock In \emph{CVPR}, 2023.

\bibitem[Zhu et~al.(2023)Zhu, Chen, Shen, Li, and Elhoseiny]{zhu2023minigpt}
Deyao Zhu, Jun Chen, Xiaoqian Shen, Xiang Li, and Mohamed Elhoseiny.
\newblock Minigpt-4: Enhancing vision-language understanding with advanced large language models.
\newblock \emph{arXiv preprint arXiv:2304.10592}, 2023.

\bibitem[Zou et~al.(2023)Zou, Dou, Yang, Gan, Li, Li, Dai, Behl, Wang, Yuan, et~al.]{zou2023generalized}
Xueyan Zou, Zi-Yi Dou, Jianwei Yang, Zhe Gan, Linjie Li, Chunyuan Li, Xiyang Dai, Harkirat Behl, Jianfeng Wang, Lu Yuan, et~al.
\newblock Generalized decoding for pixel, image, and language.
\newblock In \emph{Proceedings of the IEEE/CVF Conference on Computer Vision and Pattern Recognition}, pages 15116--15127, 2023.

\end{thebibliography}
 }

\appendix
\clearpage
\setcounter{page}{1}


\section{Implementation Details}
Our codebase is a heavily modified fork of the original ViperGPT \cite{Suris_2023_ICCV} codebase, with significantly overhauled infrastructure and several improvements for more efficient video processing, with all of \methodname\.'s modules implemented on top. We used Langchain for interfacing with large language models and including in-context examples in the prompt. For visual modules, we used BLIP-2 for Image QA, GroundingDino (with Swin-T backbone) for object detection, LaViLa for video captioning, and ByteTrack for object tracking. We evaluate most datasets with 60 frames from the video sampled uniformly over time. All model inference can be done with a single Nvidia A100 GPU, but we split the evalution over multiple GPUs for speed, especially for datasets with large test sets, like MSR-VTT. We will release our code upon publication.

\section{Datasets and Metrics Details}
\label{sec:datasets_metrics_details}
\subsection{Datasets}

\textbf{TGIF-QA \cite{tgif-cvpr2016}, MSVD-QA \cite{xu2017video}}, and \textbf{MSRVTT-QA\cite{xu2017video}} are open-ended VideoQA benchmarks automatically generated from captions, with some manual annotations. Each question has a single answer that is one word or phrase. TGIF-QA consists of GIFs that are a few seconds long, while MSVD and MSR-VTT can be up to 15 seconds. For MSVD and MSRVTT, questions are split into five categories: who, what, when, where and how, while TGIF is split into color, object, location, and number.

\noindent\textbf{iVQA} \cite{yang2021justask} is a recent open-ended benchmark based on instructional videos. It only includes visual questions, and for each answer has multiple correct answers, reducing ambiguity for each question. 

\noindent\textbf{ActivityNet-QA} \cite{yu2019activityqa} is an open-ended benchmark containing longer videos, up to 3 minutes long. It contains nine categories: motion, spatial, temporal, yes/no, color, object, location, number and `free'. 

\noindent\textbf{TVQA} \cite{lei2018tvqa} is a multiple-choice video QA dataset focused on multimodal understanding from clips of popular TV shows. Each question has 5 answers, and each question also has a ground-truth scene transcript. Questions are either visual or focused on the narrative aspect of the scene.

\begin{table*}[!htbp]
    \centering
    \begin{tabular}{lcccccccccc}
    \toprule
    \textit{ActivityNet-QA}   &  Motion & Spatial & Temporal & Yes/No & Color & Object & Location & Number & Other & Full\\
    \midrule
    Just Ask \cite{yang2021justask} & 2.3 & 1.1 & 0.3 & 36.3 & 11.3 & 4.1 & 6.5 & 0.2 & 4.7 & 12.3\\
    FrozenBiLM \cite{yang2022frozenbilm} &12.7 & \textbf{6.8} & 1.6 & 53.2 & 16.5 &17.9 &18.1 &26.2 &25.8 &25.9 \\
    \midrule
    \methodname\. (Ours)& \textbf{38.3} & 5.6 & \textbf{3.3} & \textbf{70.4} & \textbf{35.7} & \textbf{20.0} & \textbf{35.7} & \textbf{39.1} & \textbf{43.7}  & \textbf{41.3}\\
    \bottomrule
    \end{tabular}
    \caption{Zero-shot QA results per category on the ActivityNet-QA dataset.}
    \label{tab:activitynet_breakdown}
\end{table*}
\begin{table*}[!htbp]
\centering
\begin{tabular}{lccccc|lccc}
\toprule
   &  Color & Number & Location & Object & Full  & & Causal & Temporal & Full\\
\midrule
FrozenBiLM \cite{yang2022frozenbilm} & 31.3 & 67.8 & 38.2 & 40.1 & 41.9 & ViperGPT \cite{Suris_2023_ICCV} & 49.8 & 56.4 & 60.0\\
\midrule
\methodname\ (Ours) & 74.6 & 81.2 & 51.2 & 47.8 & 66.1 & \methodname\. (Ours) & 55.6 & 60.1 &  63.8\\
\bottomrule
\end{tabular}
\caption{Zero-shot video QA results per category on the TGIF-QA dataset and NeXT-QA datasets..}
\label{tab:tgif_breakdown}
\end{table*}
\begin{table*}[!htbp]
    \centering
    \begin{tabular}{clccccccc}
    \toprule
       &  & What & Who & Number & Color & When & Where & Full \\
    \midrule
    \multirow{3}{*}{\rotatebox{90}{\textbf{MSVD}}} 
    & Just Ask \cite{yang2021justask} & 7.8 & 1.7 & 74.3 & 18.8 & 3.5 & 0.0 & 13.3\\
    & FrozenBiLM \cite{yang2022frozenbilm} & 26.0 & \textbf{45.0} & 69.9 & 56.3 & 5.2 & 17.9 & 33.8 \\
    \midrule
    & \methodname\. (Ours) & \textbf{38.1} & 40.1 & \textbf{70.9} & \textbf{56.3} & \textbf{36.5} & \textbf{50.0} & \textbf{37.5} \\
    \midrule
    \multirow{3}{*}{\rotatebox{90}{\textbf{\small MSRVTT}}} 
    & Just Ask \cite{yang2021justask} & 1.8 & 0.7 & 66.3 & 0.6 & 0.6 & 4.5 & 5.6 \\
    & FrozenBiLM \cite{yang2022frozenbilm} & 10.7 & \textbf{28.7} & 55.0 & 11.4 & 9.2 & 9.3 & 16.9 \\
    \midrule
    & \methodname\. (Ours) & \textbf{14.6} & 28.1 & \textbf{67.1} & \textbf{19.3} & \textbf{22.5} & \textbf{22.6} & \textbf{22.1}  \\
    \bottomrule
    \end{tabular}
    \caption{Zero-shot QA results on the MSVD and MSR-VTT QA datasets, sepaarated by category.}
    \label{tab:msrvtt_msvd}
\end{table*}

\noindent\textbf{EgoSchema} \cite{mangalam2023egoschema} is a recent multiple-choice zero-shot benchmark focused on long-term video understanding. Videos are sourced from Ego4D and are all 3 minutes long, which questions requiring long-horizon understanding. It contains 5K video in a held-out test split and no training data. The questions and answers are created from processing Ego4D ground-truth narrations with an LLM.

\noindent\textbf{NeXT-QA}\cite{xiao2021next} is a multiple-choice benchmark designed to test causal and temporal reasoning. The answer to the question is typically found in a short timespan, while videos can be up to a minute long. Each question has 5 distinct choices.

\subsection{Metrics}
A known issue with open-ended video QA datasets is that top-1 accuracy requires a single correct answer, even though the questions are often ill-posed and have multiple valid answers.  One line of work \cite{song2023moviechat, Maaz2023VideoChatGPT} that focuses on video conversational assistants propose a different metric based on using LLMs for evaluation. In particular, they compare a sentence or paragraph output from their model to the ground-truth answer (a single word or phrase) and prompt GPT3.5 to output a binary correctness score as well as a subjective score to rate its conversational ability. While this is suitable for measuring conversational ability, we found that this metric is unreliable for measuring answer accuracy and often leads to incorrect evaluation. We thus stay with top-1 accuracy for the most fair comparison, and omit these works from our analysis.

\subsection{Per-dataset Results}
We provide breakdowns of the QA results on each dataset by question type. In Table \ref{tab:activitynet_breakdown}, we present the ActivityNet results, and in Tables \ref{tab:msrvtt_msvd} we show both the MSVD and MSR-VTT results on each category. We also include results on NeXT-QA and TGIF-QA in Table \ref{tab:tgif_breakdown}.

\begin{table*}[!htbp]
\centering
\begin{tabular}{lccccc}
\toprule
   &  TGIF-QA & MSVD & MSR-VTT & ActivityNet & iVQA \\
\midrule
\texttt{word2vec}\cite{mikolov2013efficient} &63.3 & 36.7& 19.3& 41.1& 47.5\\
BERT \cite{devlin2018bert} &61.1& \textbf{37.7}& 14.4& 39.3 & 44.3\\
FastText \cite{joulin2017bag} & \textbf{66.1} & 37.5& \textbf{22.1}& \textbf{42.3}& \textbf{50.7} \\
\midrule
No Constraint & 61.4 & 37.3 & 12.9 & 35.3 & 50.1 \\
Top-1000 & 63.8 & 37.1 & 17.8& 41.1& \textbf{50.7} \\
Type-based & \textbf{66.1} & \textbf{37.5}& \textbf{22.1}& \textbf{42.3}& - \\
\bottomrule
\end{tabular}
\caption{Ablation of components related to the output vocabularies on open-ended video dataset benchmarks. The top half of the table shows the impact of using different word-matching embeddings, and the bottom half shows the effect of restricting the output vocabulary in different ways.}
\label{tab:vocab_ablation}
\end{table*}

\section{Additional Ablations}
\label{sec:further_ablations}

\noindent\textbf{Word Matcher} Open-ended benchmarks are formulated as $K$-way classification problems, with $K$, the size of the answer vocabulary, often ranging into the thousands. For end-to-end models typically finetune pretrained features on question-answering datasets, and pick the answer from the output vocabulary with the highest score. Since our setup uses a video-language model, we map outputs to the semantically nearest word or phrase in the output vocabulary. Without this component, any prediction that is not in the vocabulary will automatically be treated as incorrect, leading to significantly worse performance on QA benchmarks. We used FastText, but other options, such as BERT or \texttt{word2vec}, could be used as well. We demonstrate the accuracy from using each of these embeddings in Table \ref{tab:vocab_ablation}. 

\noindent\textbf{Output Vocabulary} Some open-ended benchmarks have enormous answer vocabularies: MSR-VTT has 73K questions in the test set, with over 10,000 unique answers. Standard practice \cite{zellers2022merlot, yang2021justask, yang2022frozenbilm} is to use the most common 1000 answers from the training set as the vocabulary. One other input at test time is the question category: we used this to constrain the vocabulary further. For a question type $Q$, the output vocabulary is the list of all answers for questions of type $Q$ in the training set that are also among the top 1000 answers. Since our method needs no training, dynamically altering the output vocabulary is straightforward, and we found that this can significantly boost performance on datasets with poor label quality and ambiguous phrasing, such as MSRVTT-QA. The results of ablating on these components are in Table \ref{tab:vocab_ablation}. We see that the vocabulary size matters most for lower-quality labeled datasets, such as ActivityNet-QA and MSRVTT-QA, while the effect is minimal in higher-quality labels like TGIF and iVQA.

\section{Prompt}
\label{sec:prompt}
We include the full prompt containing the visual API for our model on the next page. The prompt slightly changes for each dataset: following \cite{Suris_2023_ICCV}, we exclude certain methods when running on datasets where they are not applicable. For example, we only include the \texttt{get\_summary} method on the EgoSchema benchmark. In addition to the following prompt, we include in-context examples for each dataset, which are appended to the API prompt along with the input question or command.
\begin{figure*}[b]
\begin{lstlisting}[language=Python, xleftmargin=0pt, xrightmargin=0pt]
def get_max_key(responses: Dict[str, int]) -> str
    """
    Given a dict, returns the key with the highest count.
    """

class VideoClip:
    """A Python class containing a set of frames and methods for querying them.
    Attributes
    ----------
    video : torch.Tensor
        A tensor of image frames.
    start : int
        An int describing the starting frame in this video segment.
    end : int
        An int describing the ending frame in this video segment .
    num_frames->int
        An int containing the number of frames in the video segment.
    trimmed_video: torch.Tensor
        A trimmed video from start to end of the original input tensor.
    """

    def __init__(self, video: torch.Tensor, start: int = None, end: int = None, parent_start=0, queues=None):
        """Initializes a VideoClip object.

        Parameters
        -------
        video : torch.Tensor
            A tensor of the original video.
        start : int
            An int describing the starting frame in this video segment.
        end : int
            An int describing the ending frame in this video segment.
        """

    def filter_property(self, property:str) -> VideoClip:
        """
        Given a Yes/no query, returns a VideoClip composed only of the frames
        where that statement is true.
        Parameters
        -----------
        property: str
            A query to filter the video segment with. 

        Returns: VideoClip
            A VideoClip composed only of the frames where the input
            property is true.

        Examples
        ----------
        question: What is the party for?
        def answer_question(video, possible_answers):
            party_segment = video.filter_property("Is a party happening?")
            responses = vid_segment.video_query("What is the party for?", possible_answers)
            return get_max_key(responses)
        """

    def filter_object(self, object: str) -> VideoClip:
        """
        Given a object, returns a VideoClip composed only of frames where
        that object is present. 
        Parameters
        -----------
        object: str
            The object to look for.

        Returns: VideoClip
            A VideoClip composed only of the frames containing the input 
            object.

        Examples
        ----------
        question: What color is the skier's jacket?
        def answer_question(video, possible_answers):
            skier_clip = video.filter_object("skier")
            skier_boxes = video.find("skier")
            jacket_boxes = skier_clip.find("jacket")
            responses = jacket_boxes.video_query("What color is this jacket?", possible_answers)
            return get_max_key(responses)
        """
\end{lstlisting}
\end{figure*}
\newpage    
\begin{figure*}[b]
\begin{lstlisting}[language=Python, xleftmargin=0pt, xrightmargin=0pt,firstnumber=81]
       def video_query(self, query: str, possible_answers: List[str]) -> Dict:
        """Answers a query for each frame in the video and returns a dict with the count of responses.
        Parameters
        -----------
        query: str
            The question to be answered.
        possible_answers : List[str]
            The list of possible answers for output.

        Returns: Dict
            The query answers, grouped by how many frames they occur for.

        Examples
        --------
        question: what is the person doing?
        def answer_question(video, possible_answers):
            responses = video.video_query("What is the person doing?", possible_answers)
            return get_max_key(responses)
        """

    def get_caption(self, index: int) -> str:
        """
        Gets a caption of the frame at that index in the video segment. 
        Parameters
        ----------
        index: int  
            The index of the frame to use. Range is [0, self.num_frames -1].

        Returns : str
            The image caption of the frame at that index.
        """

    def find(self, object: str) -> VideoClip:
        """
        Finds all bounding boxes around a certain object in a video segment, 
        and collates them into a collection of frames.

        Parameters
        ---------
        object: str
            The object to look for. 

        Returns : VideoClip
            A VideoClip object composed of crops of the object.
        """

    # This is only included in the prompt if we can get the script.
    def get_script(self) -> str:
        """
        Returns: 
            A string script of the speech spoken during the video, if available.
        """
    
    # This is only included in the prompt for the Egoschema evaluation, 
    # and should only be used if a sufficient video captioning model exists.
    def get_summary(self) -> str:
        """
        Returns: str
            A string summary representing the narrative of the video.
        """

    def track_objects(self, input_boxes: List[torch.Tensor]): -> List(STrack)
        """
        Runs a tracker on a set of input bounding boxes, representing some object(s)
        detected over time. Returns the boxes grouped by track ID. 

        Parameters
        ---------
        input_boxes: List[torch.Tensor]
            A list of all the detected boxes at each frame in the video. 

        Returns: List[STrack]
            A list of tracked objects with bounding box and time information.
        --------
        """
\end{lstlisting}
\end{figure*}
\begin{figure*}[htbp]
\begin{lstlisting}[language=Python, xleftmargin=0pt, xrightmargin=0pt, firstnumber=156]
    def choose_option(self, question:str, context: Dict, options: List[str]) -> str:
        """
        Uses a language model to choose the option that best answers the question
        given the input context. 

        Parameters
        ----------
        question: str   
            The input query. 
        context: Dict
            Any useful context, such as scripts, visual information, or summaries.
        options: List[str]
            The list of options to choose from, numbered.
        
        Returns: str
            A string detailing which number option was chosen with reasoning.

        Examples
        ---------
        question: How was the toy bear moved to the front?
        def answer_question(video, possible_answers):
            vid_seg = video.trim(0, len(video) // 4) # consider the star
            bear_seg = vid_seg.filter_object("bear")
            image_context = bear_seg.get_caption(bear_seg.num_frames // 2)
            activity_context = bear_seg.video_query("What is this?")
            context = {"caption": image_context, "activity": activity_context}
            answer = bear_seg.choose_option("how was the toy bear moved to the front?", context, possible_answers)
            return answer
        """















































\end{lstlisting}
\end{figure*}

\end{document}